\documentclass[conference]{IEEEtran}
\IEEEoverridecommandlockouts
\usepackage[T1]{fontenc}
\usepackage[utf8]{inputenc}

\usepackage{amsmath,amssymb,amsfonts}

\usepackage{graphicx}
\usepackage{subcaption}

\usepackage{booktabs}
\usepackage{array}
\usepackage{tabularx}
\newcolumntype{Y}{>{\raggedright\arraybackslash}X}
\newcolumntype{L}[1]{>{\raggedright\arraybackslash}p{#1}}

\usepackage{algorithm}
\usepackage{algpseudocode}

\usepackage{listings}

\usepackage{textcomp}
\usepackage{xcolor}
\usepackage{nicefrac}

\usepackage{xurl}

\usepackage{microtype}

\usepackage{hyperref}
\graphicspath{{./images/}}

\def\BibTeX{{\rm B\kern-.05em{\sc i\kern-.025em b}\kern-.08em
T\kern-.1667em\lower.7ex\hbox{E}\kern-.125emX}}

\title{DAH-Net: A Dual-Attention Hybrid Network for Interpretable and Robust EEG-Based Emotion Recognition}

\author{\IEEEauthorblockN{1\textsuperscript{st} S M Rakib Ul Karim}
\IEEEauthorblockA{\textit{Department of Electrical \& Computer Engineering} \\
\textit{University of Missouri}\\
Columbia, Missouri, USA \\
skarim@missouri.edu}
\and
\IEEEauthorblockN{2\textsuperscript{nd} Diponkor Bala}
\IEEEauthorblockA{\textit{Department of Computer Science and Engineering} \\
\textit{City University}\\
Savar, Dhaka-1340, Bangladesh \\
diponkor.b@gmail.com}
\and
\IEEEauthorblockN{3\textsuperscript{rd} Wenyi Lu}
\IEEEauthorblockA{\textit{Department of Computer Science} \\
\textit{University of Missouri}\\
Columbia, Missouri, USA \\
wldh6@mail.missouri.edu}
\and
\IEEEauthorblockN{4\textsuperscript{th} Rownak Ara Rasul}
\IEEEauthorblockA{\textit{Department of Computer Science} \\
\textit{University of Missouri}\\
Columbia, Missouri, USA \\
rrasul@missouri.edu}
\and
\IEEEauthorblockN{5\textsuperscript{th} Sean Goggins}
\IEEEauthorblockA{\textit{Department of Electrical \& Computer Engineering} \\
\textit{University of Missouri}\\
Columbia, Missouri, USA \\
gogginss@missouri.edu}
}

\begin{document}

\maketitle
\thispagestyle{empty}
\pagestyle{empty}

%%%%%%%%%%%%%%%%%%%%%%%%%%%%
\begin{abstract}
EEG-based emotion recognition supports affective brain-computer interfaces and mental health monitoring, yet remains challenged by signal complexity, subject variability, and limited interpretability. We propose DAH-Net, a dual-attention hybrid network integrating 1D-CNN, BiLSTM, and dual multi-head attention (16+8 heads) for three-class EEG emotion classification. Evaluated on 2,479 samples with 988 EEG features, DAH-Net achieves 99.19\% held-out test accuracy with a 0.81\% train-test gap, outperforming RF (96.17\%), SVM (96.77\%), MLP (97.18\%), and Transformer (98.19\%) baselines. Friedman testing ($\chi^2=28.54$, $p<0.001$) and post-hoc Wilcoxon comparisons confirm statistical significance. Feature-level analysis using Random Forest importance, SHAP attribution, and feature-category isolation shows that covariance features achieve near-baseline standalone accuracy (94.96\%), while eigenvalue features show limited standalone performance (84.07\%) but provide compact complementary information. The compact architecture (3.33M parameters, approximately 13.3MB using 32-bit weights) suggests potential for future lightweight EEG-based affective computing, pending subject-independent and external validation.
\end{abstract}

%%%%%%%%%%%%%%%%%%%%%%%%%%%%

%%%%%%%%%%%%%%%%%%%%%%%%%%%%
\section{Introduction}
Electroencephalography (EEG) enables non-invasive, high-temporal-resolution decoding of neural activity and is widely explored for emotion recognition in affective computing, mental health assessment, and adaptive human-computer interfaces \cite{picard2000affective, muhl2014eeg}. Distinct oscillatory and connectivity patterns reflect emotional states such as neutral, positive, and negative \cite{zheng2015investigating, alarcao2017emotions}, offering opportunities for objective affect monitoring. However, practical deployment remains challenging due to high dimensionality, subject variability, and non-stationarity in EEG signals \cite{craik2019deep}.

Early pipelines relied on handcrafted statistical, spectral, and connectivity features with SVMs and Random Forests \cite{koelstra2011deap, zheng2015investigating}, achieving moderate performance while requiring extensive feature engineering. These remain valuable baselines for understanding performance-cost tradeoffs \cite{tangermann2012review}. Deep learning has improved decoding through CNNs \cite{schirrmeister2017deep}, BiLSTMs \cite{bashivan2015learning, tao2020eeg}, and transformer attention for long-range dependencies \cite{vaswani2017attention, wu2020transfer}. Despite strong benchmark results from attention-based models \cite{song2018eeg, zhong2020eeg, li2019regional}, gaps remain in multi-stage attention design, systematic feature analysis, and rigorous statistical benchmarking. To address these gaps, we investigate the following research questions (RQs):

\begin{itemize}
    \item \textbf{RQ1:} To what extent do hybrid deep learning architectures that integrate convolutional, recurrent, and attention-based components improve EEG-based emotion classification performance, robustness to overfitting, and interpretability compared to classical and deep learning baselines?
    \item \textbf{RQ2:} What are the most critical EEG feature categories for emotion recognition, and how do they contribute to model performance through feature-category isolation and multi-method importance analysis?
    \item \textbf{RQ3:} Can hybrid combinations of classical machine learning and neural network approaches improve EEG emotion classification through ensemble methods and voting strategies?
\end{itemize}

We propose DAH-Net, integrating residual 1D-CNN, BiLSTM, and dual multi-head self-attention with advanced regularization (dropout, layer normalization, label smoothing, weight decay) for improved within-dataset EEG emotion recognition. Feature-level analysis uses Random Forest importance, SHAP attribution, and feature-category isolation to identify discriminative EEG patterns across statistical, frequency, covariance, and eigenvalue categories. Statistical validation via Friedman tests and post-hoc Wilcoxon comparisons assesses reliability against baseline models.

%%%%%%%%%%%%%%%%%%%%%%%%%%%%

\section{Related Work}

\textbf{Classical and Early Deep Learning Approaches.} EEG-based emotion recognition has traditionally relied on handcrafted feature engineering combined with classical machine learning algorithms. Early work \cite{koelstra2011deap, zheng2015investigating} achieved 70–85\% accuracy on the DEAP dataset using manual statistical (mean, variance, skewness) and frequency-domain features (power spectral density) paired with Support Vector Machines or Random Forests. These approaches required extensive domain expertise and suffered from limited representational capacity \cite{craik2019deep}. CNN-based architectures subsequently achieved 89–92\% accuracy \cite{tripathi2017using, li2018hierarchical}, enabling end-to-end learning from preprocessed EEG representations \cite{schirrmeister2017deep}.

\textbf{Recurrent and Attention-Based Architectures.} BiLSTM networks capture temporal dependencies in EEG sequences (89–90\% accuracy) \cite{alhagry2017emotion, bashivan2015learning}. Transformer-based attention mechanisms model long-range feature interactions. Vision Transformers and self-attention modules achieved 87–93.8\% accuracy \cite{song2018eeg, tao2020eeg, zhong2020eeg}, leveraging multi-head attention to weight feature relevance dynamically. Attention mechanisms have proven particularly effective for affective computing, enabling both improved accuracy and interpretability through attention weight visualization \cite{wu2020transfer, lu2022transformer}.

\textbf{Hybrid Architectures and Feature Analysis.} Recent work has combined CNNs, BiLSTMs, and attention mechanisms to achieve complementary spatial, temporal, and long-range modeling. Hybrid CNN-BiLSTM-Attention architectures have reached 93–97\% accuracy on DEAP benchmarks \cite{li2019regional, iacono2024multi}, suggesting that integrating multiple modalities improves robustness. However, gaps remain in multi-stage attention design, systematic EEG feature category analysis, rigorous statistical benchmarking \cite{demsar2006statistical}, and detailed feature-level interpretability combining SHAP, importance rankings, and ablation studies \cite{shapley2016value}.

\textbf{This Work.} We address these gaps by proposing DAH-Net, a dual-attention hybrid architecture with advanced regularization and comprehensive feature-level analysis. Our contributions include: (1) a dual multi-head attention design (16+8 heads) with residual CNN blocks and BiLSTM temporal modeling; (2) systematic feature-category isolation and multi-method importance analysis characterizing the discriminative contribution of each EEG feature category; (3) rigorous statistical validation with Friedman tests and Wilcoxon comparisons; and (4) reproducible evaluation on a 2,479-sample public dataset with transparent code and hyperparameters.

%%%%%%%%%%%%%%%%%%%%%%%%%%%%

\section{Methodology}

\subsection{Dataset Description}

\begin{figure*}[t]
\centering
\includegraphics[width=.8\textwidth]{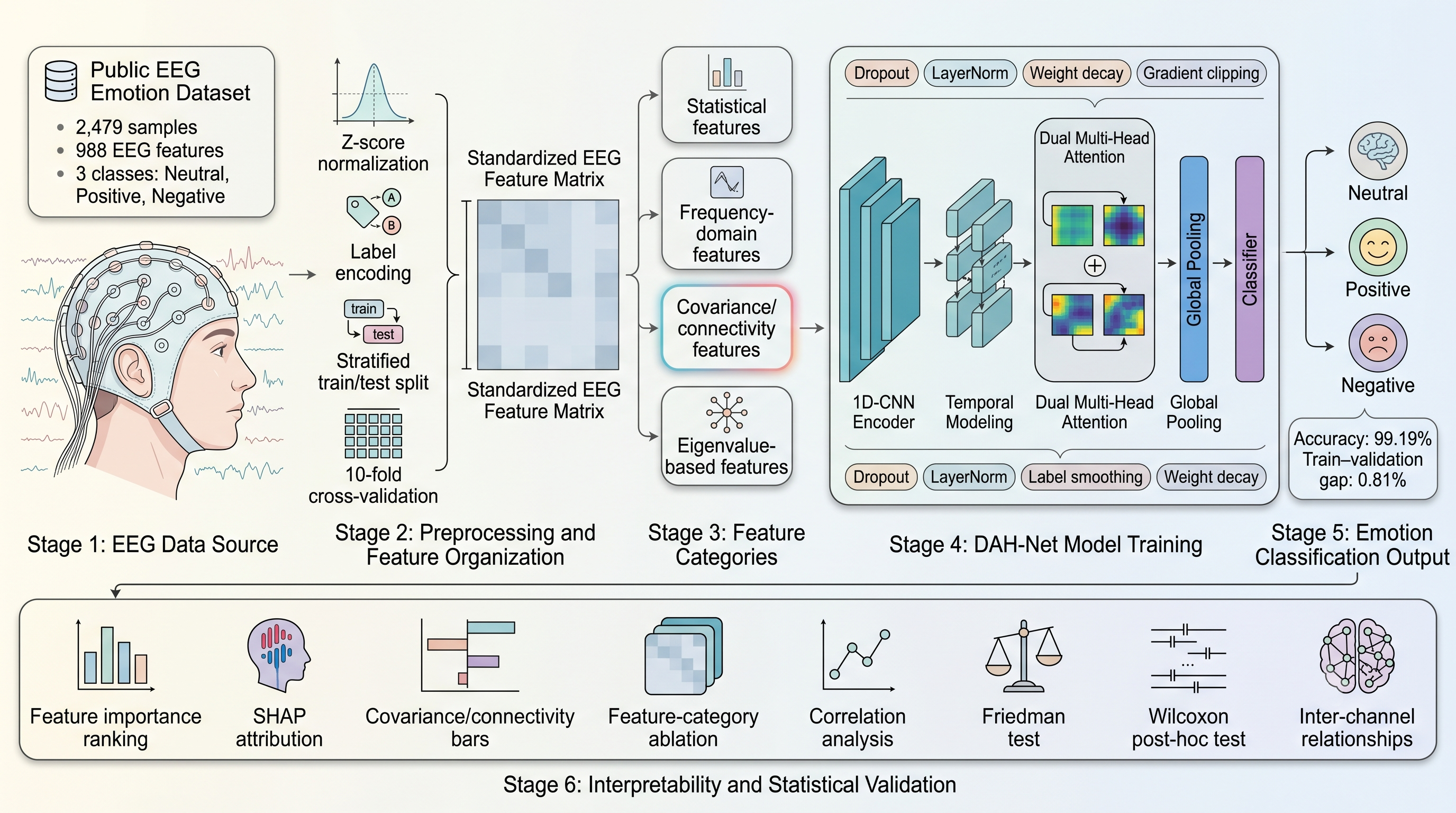}
\caption{\textbf{Overview of the proposed EEG emotion classification workflow.} The pipeline begins with a publicly available EEG emotion dataset containing 2,479 samples, 988 extracted EEG features, and three emotional states: neutral, positive, and negative. After preprocessing, including normalization, label encoding, stratified data partitioning, and cross-validation, the standardized feature representations are grouped into statistical, frequency-domain, covariance/connectivity, and eigenvalue-based categories. The proposed DAH-Net then learns discriminative affective representations using residual 1D convolutional feature extraction, BiLSTM-based temporal modeling, dual multi-head self-attention, global pooling, and a fully connected classifier.}
\label{fig:Overview}
\end{figure*}

This study employs a publicly available EEG emotion classification dataset originally curated for mental state analysis \cite{bird2019mental}. The dataset consists of EEG-derived feature representations corresponding to three emotional states: neutral, positive, and negative, commonly adopted in prior affective computing research \cite{koelstra2011deap, zheng2015investigating}. To support transparency and reproducibility, the complete dataset used in this study is publicly accessible via Kaggle.\footnote{\href{https://www.kaggle.com/datasets/birdy654/eeg-brainwave-dataset-mental-state?resource=download}{Kaggle EEG Brainwave Dataset for Mental State}.}
The dataset contains $N = 2{,}479$ samples, each representing a short EEG recording segment captured from a single participant during an emotionally evoked mental state; each segment is summarized by a $D = 988$-dimensional feature vector extracted from the preprocessed EEG signals \cite{bird2019mental}. These features encompass multiple complementary categories, including statistical descriptors (e.g., mean, variance, skewness, kurtosis), frequency-domain measures (e.g., power spectral density within alpha, beta, and gamma bands), covariance-based connectivity features that capture inter-channel relationships \cite{duan2013differential}, and eigenvalue-based measures reflecting dominant oscillatory modes. Each sample is denoted as $\mathbf{x}_i \in \mathbb{R}^{988}$ and associated with a categorical label $y_i \in \{0, 1, 2\}$ corresponding to neutral, positive, and negative emotional states, respectively. The dataset exhibits an approximately balanced class distribution, with each emotion category accounting for roughly one-third of the samples. The complete classification workflow, from raw EEG 
features through preprocessing, feature grouping, and DAH-Net inference, is illustrated in Figure~\ref{fig:Overview}.

\subsection{Data Preprocessing and Normalization}

Z-score normalization standardized features to zero mean and unit variance to prevent high-magnitude features from dominating optimization \cite{goodfellow2016deep}. Data was split 80/20 (train/test) using stratified sampling to preserve class proportions and prevent information leakage \cite{kaufman2012leakage}.

\subsection{Model Architecture and Components}

The proposed Enhanced Transformer-CNN-BiLSTM (DAH-Net) architecture integrates multiple learning modalities for comprehensive EEG representation \cite{huang2023model}. The model processes 988-dimensional EEG feature vectors through a hierarchical pipeline combining convolutional spatial analysis, bidirectional temporal modeling, and multi-scale attention mechanisms \cite{li2019regional}.

\begin{figure*}[t]
\centering
\includegraphics[width=\textwidth,height=0.25\textheight,keepaspectratio]{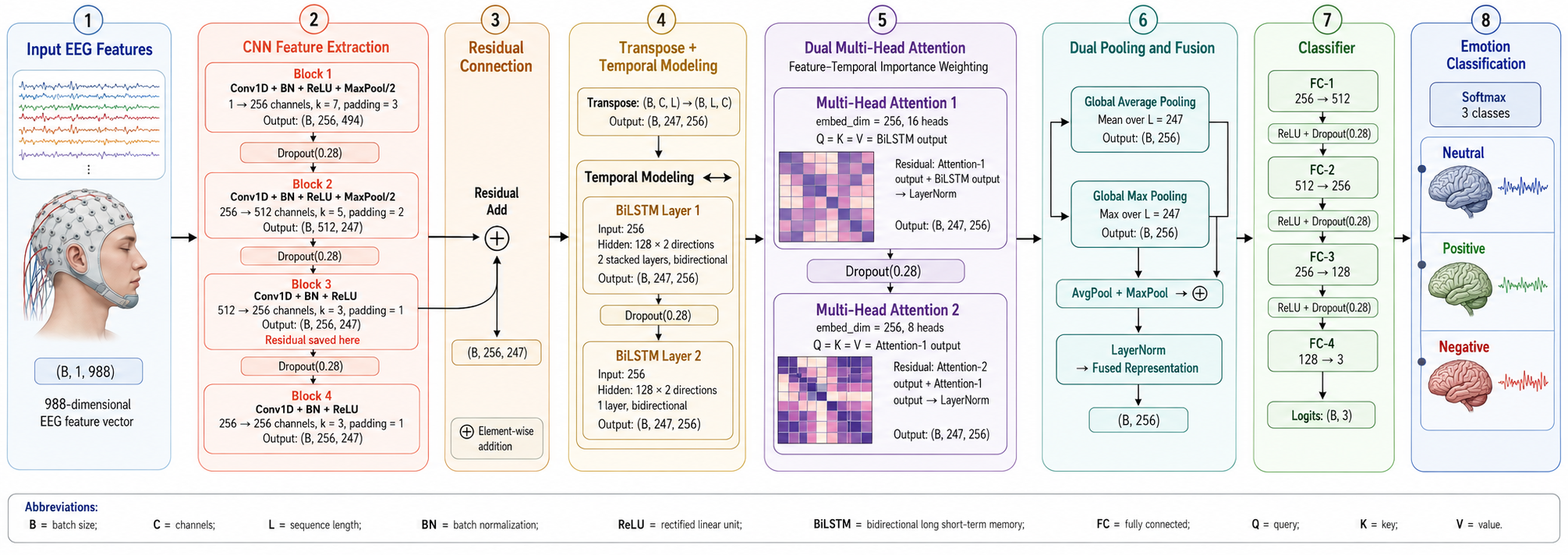}
\caption{\textbf{Architecture of the proposed DAH-Net for EEG-based emotion classification.} The model receives a 988-dimensional EEG feature representation and processes it through a residual 1D convolutional encoder to extract local discriminative patterns. The encoded features are then transposed and passed through two BiLSTM layers to capture bidirectional temporal dependencies. Dual multi-head self-attention modules further refine the learned representation by emphasizing informative feature–temporal relationships. Global average and max pooling are combined to generate a compact fused representation, which is passed through a fully connected classifier and softmax layer to predict three emotional states: neutral, positive, and negative.}
\label{fig:enhanced_model}
\end{figure*}

DAH-Net combines 1D-CNN, BiLSTM, and dual multi-head attention to model spatial-temporal EEG dynamics (Figure~\ref{fig:enhanced_model}).

Baselines: Random Forest (100 trees), Support Vector Machine (RBF kernel, $C=1.0$), Multilayer Perceptron (256/128 neurons with ReLU), and a plain Transformer encoder (multi-head self-attention with 8 heads, 2 encoder layers, dropout=0.35) providing reference points for classical, neural, and attention-based modeling, respectively.

Spatial features extracted via 1D-CNN with residual connections: $\mathbf{h}^{(l)} = \text{ReLU}(\mathbf{W}^{(l)} * \mathbf{x} + \mathbf{b}^{(l)})$. Temporal dependencies modeled by BiLSTM with forward-backward concatenation capturing past-future context. Dual multi-head attention (16+8 heads) with scaled dot-product: $\text{Attention}(\mathbf{Q}, \mathbf{K}, \mathbf{V}) = \text{softmax}(\frac{\mathbf{Q}\mathbf{K}^T}{\sqrt{d_k}})\mathbf{V}$. Dual pooling (max+average) aggregates representations, passed through softmax classifier producing 3-class probabilities.

\begin{table}[t]
\centering
\footnotesize
\caption{Enhanced Model Architecture Components}
\label{tab:architecture_spec}
\begin{tabular}{p{2.2cm}p{2.4cm}p{2.2cm}}
\toprule
\textbf{Component} & \textbf{Configuration} & \textbf{Purpose} \\
\midrule
1D-CNN Encoder & 4 conv + 1 res. & Spatial extract. \\
BiLSTM Layers & 2 bidir., 256 units & Temporal model. \\
Attention Heads & 16+8 heads & Long-range \\
Global Pooling & Max+Avg. & Aggregation \\
Dropout & 0.28 & Prevent co-adapt. \\
Output & Softmax (3) & Classification \\
\bottomrule
\end{tabular}
\end{table}

\subsection{Training Procedure}

AdamW optimizer ($\beta_1=0.9$, $\beta_2=0.999$, 
$\lambda=5\times10^{-5}$ weight decay) with cosine annealing 
warm restarts ($\eta_0=0.0008$, $T_0=50$, $T_{\text{mult}}=1$). 
Loss: cross-entropy with label smoothing ($\epsilon=0.01$). 
Regularization: dropout (0.28), gradient clipping (norm=1.0), 
early stopping (patience=75). Max 400 epochs, batch size 32. 
Best model retained from maximum validation accuracy.

\begin{table}[t]
\centering
\footnotesize
\caption{Regularization and Optimization Techniques}
\label{tab:regularization_techniques}
\begin{tabular}{p{2cm}p{2cm}p{1.8cm}}
\toprule
\textbf{Technique} & \textbf{Parameter} & \textbf{Purpose} \\
\midrule
Dropout & 0.28 & Co-adaptation control \\
Layer Norm & Per layer & Stabilize \\
Label Smooth. & $\epsilon=0.01$ & Reduce overconfidence \\
Weight Decay & $\lambda=5\times10^{-5}$ & Reduce overfit. \\
Grad. Clip & Norm=1.0 & Stable gradients \\
AdamW & $\beta_1=0.9$ & Adaptive LR \\
LR Schedule & Cosine ann. & Smooth decay \\
Data Aug. & Noise+scaling & Robustness \\
Early Stop & Patience=75 & Convergence \\
\bottomrule
\end{tabular}
\end{table}

\begin{algorithm}[t]
\caption{Training of DAH-Net for EEG Emotion Classification}
\label{alg:dahnet_training}
\begin{algorithmic}[1]
\Require EEG dataset $\mathcal{D}$; training set $\mathcal{D}_{train}$; 
         validation set $\mathcal{D}_{val}$; epochs $E=400$; batch size 
         $B=32$; learning rate $\eta_0=0.0008$; weight decay 
         $\lambda=5\times10^{-5}$; label smoothing $\epsilon=0.01$; 
         patience $P=75$
\Ensure Trained DAH-Net model with maximum validation accuracy

\State Fit Z-score scaler on $\mathcal{D}_{train}$; apply fitted scaler 
       to $\mathcal{D}_{val}$ (no leakage)
\State Initialize DAH-Net model parameters $\theta$
\State Initialize AdamW optimizer with $\eta_0$ and $\lambda$
\State Initialize cosine annealing scheduler 
       ($T_0{=}50$, $T_{mult}{=}1$)
\State $A_{best} \leftarrow 0$, $\theta_{best} \leftarrow \theta$, 
       $counter \leftarrow 0$

\For{$epoch = 1$ to $E$}
    \For{each mini-batch $(X, y)$ in $\mathcal{D}_{train}$}
        \State $z \leftarrow M_{\theta}(X)$
        \State $L_{cls} \leftarrow \mathrm{CrossEntropyLS}(z, y, \epsilon)$
        \State Backpropagate $L_{cls}$ with gradient clipping 
               ($\text{norm}{=}1.0$)
        \State Update $\theta$ using AdamW
    \EndFor

    \State Compute validation accuracy $A_{val}$ on $\mathcal{D}_{val}$

    \If{$A_{val} > A_{best}$}
        \State $A_{best} \leftarrow A_{val}$
        \State $\theta_{best} \leftarrow \theta$
        \State $counter \leftarrow 0$
    \Else
        \State $counter \leftarrow counter + 1$
    \EndIf

    \If{$counter \geq P$}
        \State \textbf{break}
    \EndIf

    \State Step cosine annealing scheduler
\EndFor

\State \Return $M_{\theta_{best}}$
\end{algorithmic}
\end{algorithm}

\subsection{Evaluation Metrics}

Model performance was evaluated using accuracy, precision, recall, and macro-F1. Final performance used a stratified 80/20 train-test split, with 10\% of the training partition reserved for validation-based early stopping and checkpoint selection; the held-out test set was accessed only once after training. For statistical comparison, 10-fold stratified cross-validation was used across five models, followed by Friedman testing and post-hoc Wilcoxon signed-rank tests with Bonferroni correction ($\alpha=0.05$, $\alpha_{\text{corrected}}=0.005$). The overfitting gap was computed as training accuracy minus validation accuracy.

%%%%%%%%%%%%%%%%%%%%%%%%%%%%

\section{Results}
\label{sec:results}

\subsection{Model Performance and Comparative Evaluation}

To answer RQ1, this subsection evaluates whether our proposed hybrid deep learning architectures improve EEG-based emotion classification performance, robustness to overfitting, and interpretability relative to classical and deep learning baselines. All model comparisons were evaluated using 10-fold stratified cross-validation, and final test performance was reported on the held-out test set. Statistical significance was examined using the Friedman test followed by post-hoc Wilcoxon signed-rank tests with Bonferroni correction ($\alpha = 0.05$).

\subsubsection{Baseline Model Performance}

\begin{figure}[!t]
\centering
\begin{subfigure}{0.45\textwidth}
\includegraphics[width=\textwidth]{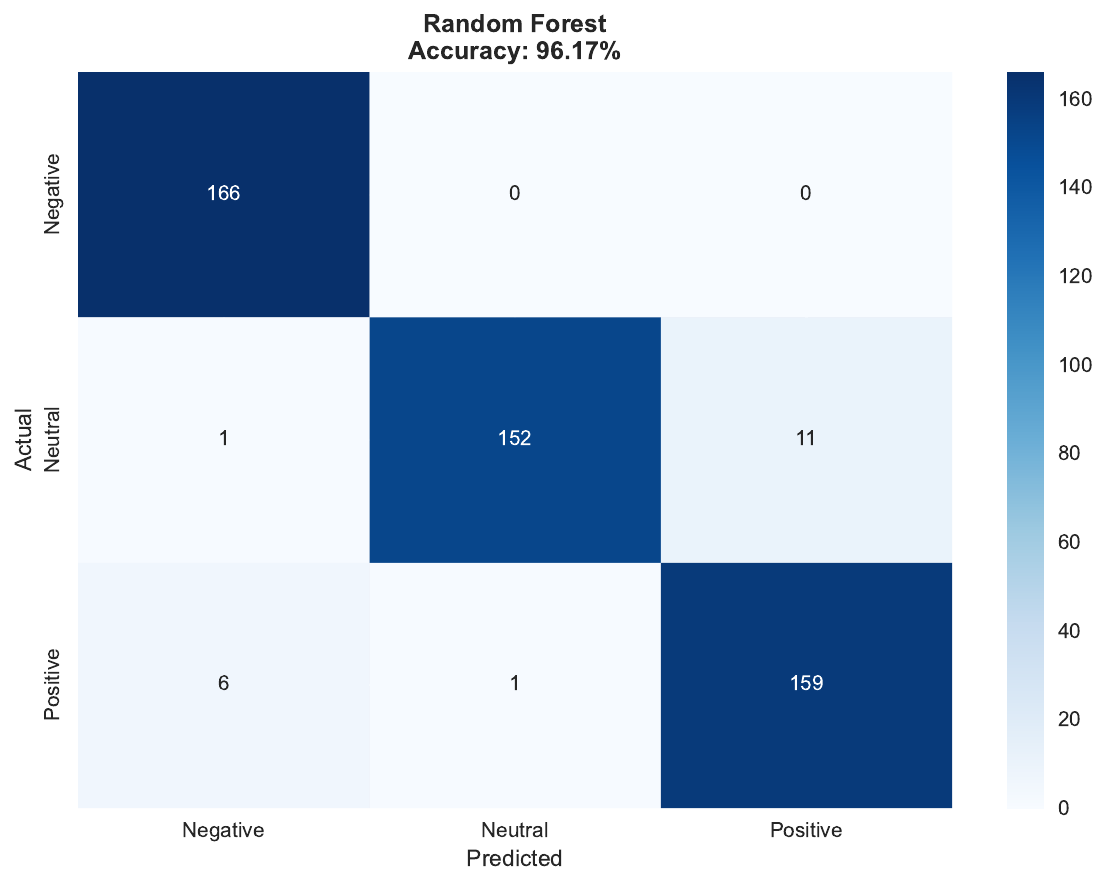}
\caption{Random Forest (96.17\% Accuracy)}
\label{fig:baseline_rf_cm}
\end{subfigure}
\hfill
\begin{subfigure}{0.45\textwidth}
\includegraphics[width=\textwidth]{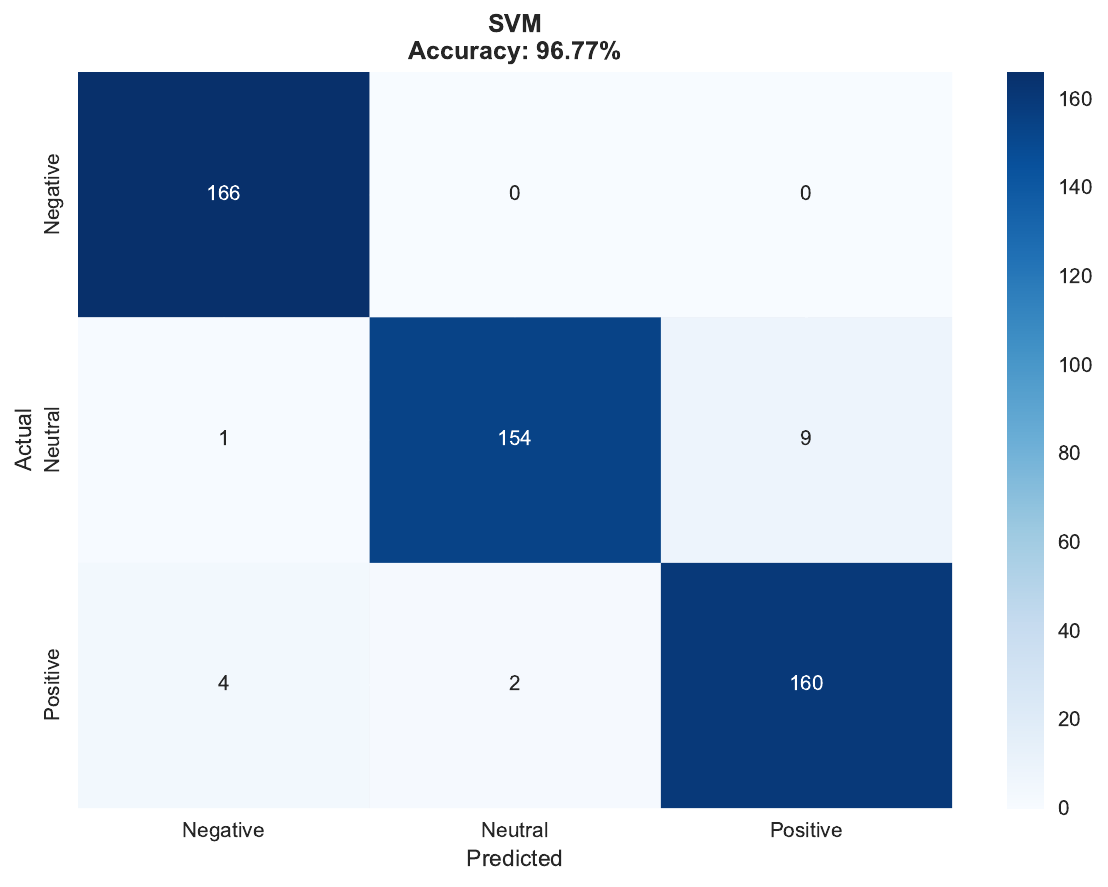}
\caption{SVM (96.77\% Accuracy)}
\label{fig:baseline_svm_cm}
\end{subfigure}
\hfill
\begin{subfigure}{0.45\textwidth}
\includegraphics[width=\textwidth]{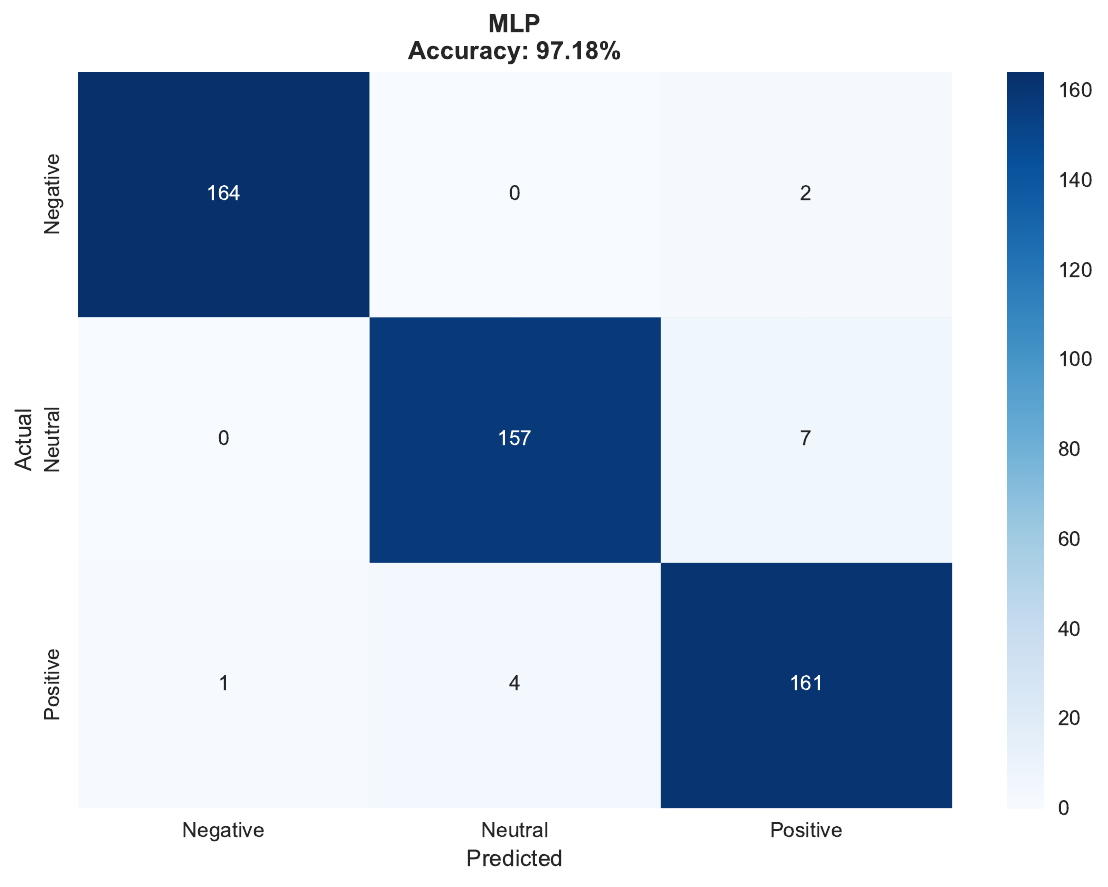}
\caption{MLP (97.18\% Accuracy)}
\label{fig:baseline_mlp_cm}
\end{subfigure}
\caption{Confusion Matrices for Baseline Models Showing Per-Class Classification Performance}
\label{fig:baseline_cm}
\end{figure}

Four baselines were evaluated on 988-dimensional EEG features: Random Forest (100 trees), SVM (RBF kernel), MLP (2 layers, 256/128 units), and a plain Transformer encoder (8 heads, 2 layers, dropout=0.35). Confusion matrices (Figure~\ref{fig:baseline_cm}) show balanced per-class performance with minimal bias. These establish strong benchmarks for hybrid deep learning comparison.

\begin{table}[t]
\centering
\caption{Model Performance on EEG Emotion Classification}
\label{tab:baseline_results}
\begin{tabular}{lcccc}
\toprule
\textbf{Model} & \textbf{Acc. (\%)} & \textbf{Prec. (\%)} & \textbf{Rec. (\%)} & \textbf{F1 (\%)} \\
\midrule
Random Forest & 96.17 & 96.30 & 96.17 & 96.17 \\
SVM           & 96.77 & 96.90 & 96.77 & 96.77 \\
MLP           & 97.18 & 97.10 & 97.18 & 97.08 \\
Transformer   & 98.19 & 98.30 & 98.19 & 98.19 \\
\textbf{DAH-Net} & \textbf{99.19} & \textbf{99.20} & \textbf{99.19} & \textbf{99.19} \\
\bottomrule
\end{tabular}
\end{table}

\subsubsection{DAH-Net Architecture Performance}

\paragraph{Training Dynamics and Convergence}

\begin{figure*}[t]
\centering
\includegraphics[width=0.88\textwidth]{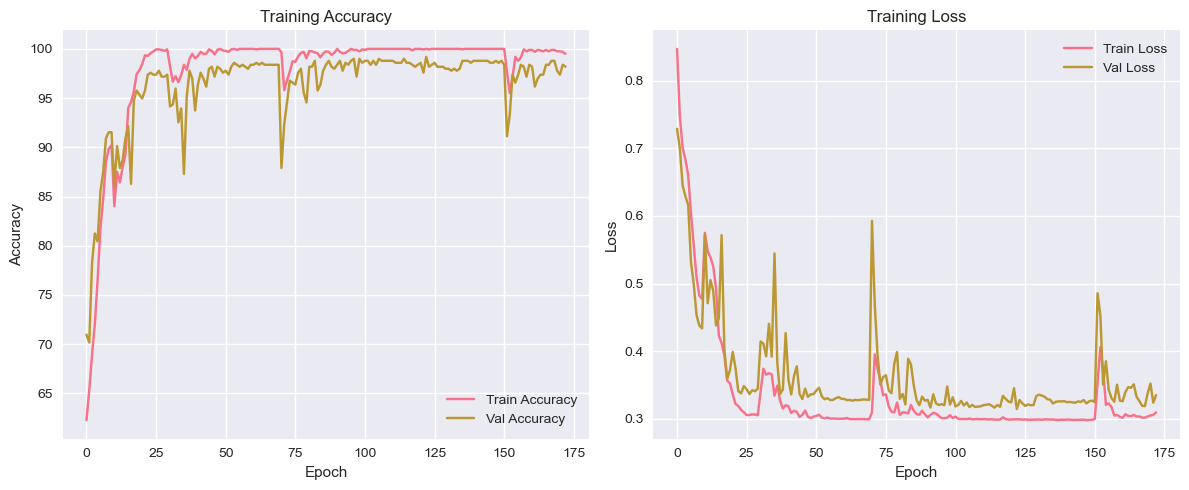}
\caption{Training and validation accuracy/loss curves for the proposed 
DAH-Net model up to the selected checkpoint. The internal validation split was used solely for monitoring convergence and early stopping during training; final test accuracy (99.19\%) was evaluated once on the held-out 20\% 
test set after training was fully complete.}
\label{fig:training_curves}
\end{figure*}

Figure~\ref{fig:training_curves} illustrates the model's learning progression, showing stable convergence with minimal overfitting. The training accuracy increases to 100.00\%, while the validation accuracy reaches 99.19\%, demonstrating strong generalization. The training and validation losses decrease substantially and stabilize at low values under label smoothing, without divergence. The small overfitting gap of 0.81\% highlights the model's robustness, achieved through dropout, label smoothing, and early stopping.

\begin{table}[t]
\centering
\small
\caption{Training Dynamics and Convergence Metrics for DAH-Net}
\label{tab:training_dynamics}
\begin{tabular}{ll}
\toprule
\textbf{Metric} & \textbf{Value} \\
\midrule
Total Epochs        & 400 (max) \\
Best Epoch          & 212 \\
Train Accuracy      & 100.00\% \\
Best Val. Accuracy\textsuperscript{$\dagger$} & 99.19\% \\
Train Loss          & Stabilized at low value \\
Val. Loss           & Stabilized at low value \\
Overfitting Gap     & 0.81\% \\
Time to Conv.       & $\sim$2 hrs \\
Early Stop Pat.     & 75 epochs \\
Stability           & No divergence \\
\bottomrule
\multicolumn{2}{l}{\small\textsuperscript{$\dagger$}Used for checkpoint selection; final held-out test}\\
\multicolumn{2}{l}{\small accuracy was evaluated once after training.}\\
\end{tabular}
\end{table}

\paragraph{Overall Performance}

The hybrid model achieved 99.19\% test accuracy with only 0.81\% train-test gap, substantially outperforming all baselines (RF 96.17\%, SVM 96.77\%, MLP 97.18\%), demonstrating effective overfitting control and robust generalization.

\paragraph{Per-Class Performance}

Table~\ref{tab:per_class_performance} presents a detailed breakdown of per-class performance. The Negative class achieves perfect precision, recall, and F1 (100.00\%), while Neutral and Positive exceed 98.17\% on all per-class metrics. Macro-averaged precision reaches 99.20\%, with macro recall and F1 at 99.19\%. Performance variation is minimal (range: 98.17\%-100.00\%), indicating balanced discrimination across all three emotion categories, an important property for downstream affective monitoring where class-specific misclassification costs may differ.

\begin{table}[t]
\centering
\caption{Per-Class Performance Breakdown of DAH-Net on the Held-Out Test Set 
(N\,=\,496, 20\% stratified split, evaluated once after training completion)}
\label{tab:per_class_performance}
\begin{tabular}{lcccc}
\toprule
\textbf{Class} & \textbf{Prec.~(\%)} & \textbf{Rec.~(\%)} & \textbf{F1~(\%)} & \textbf{Support} \\
\midrule
Negative & 100.00 & 100.00 & 100.00 & 166 \\
Neutral  &  99.38 &  98.17 &  98.77 & 164 \\
Positive &  98.21 &  99.40 &  98.80 & 166 \\
\midrule
\textbf{Macro Avg.}    & \textbf{99.20} & \textbf{99.19} & \textbf{99.19} & \textbf{496} \\
\textbf{Weighted Avg.} & \textbf{99.20} & \textbf{99.19} & \textbf{99.19} & \textbf{496} \\
\bottomrule
\end{tabular}
\end{table}

\paragraph{Confusion Matrix and Error Analysis}

\begin{figure}[!t]
\centering
\includegraphics[width=0.98\columnwidth]{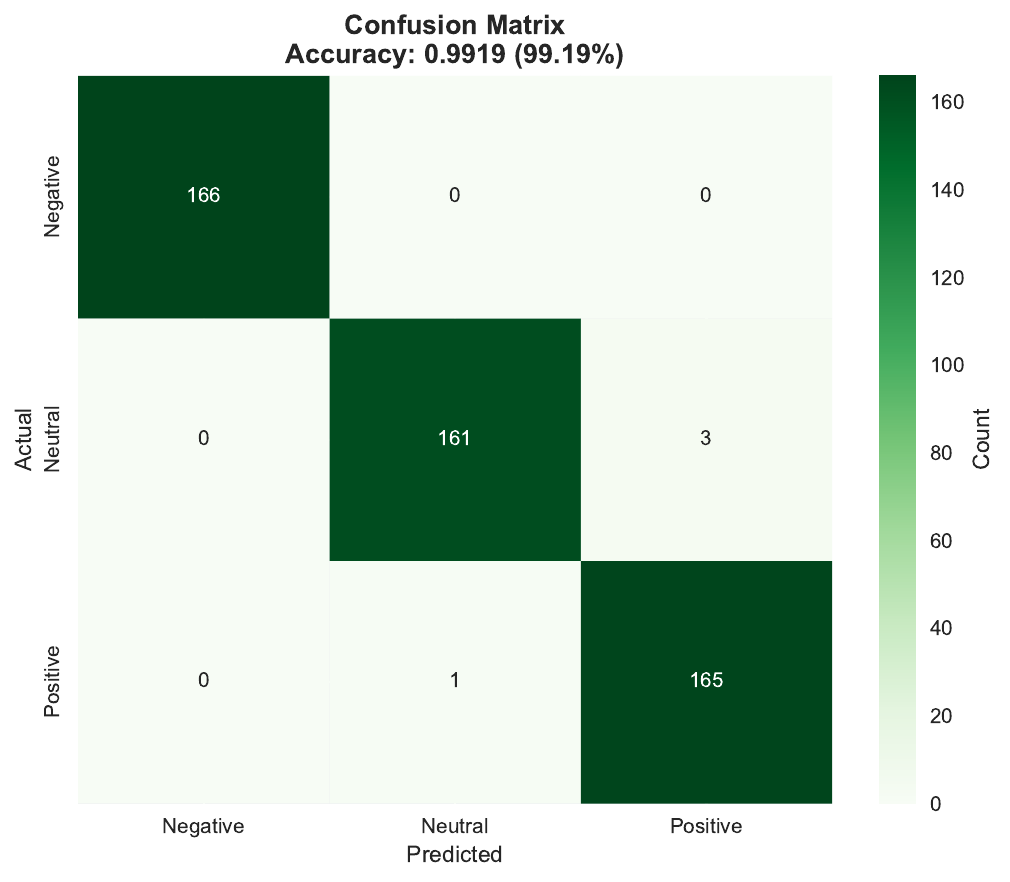}
\caption{Enhanced Model Confusion Matrix (99.19\% Accuracy)}  
\label{fig:enhanced_cm}
\end{figure}

Figure~\ref{fig:enhanced_cm} shows the confusion matrix with strong diagonal dominance: Negative achieves 100.00\% recall (166/166), Positive 99.40\% (165/166), and Neutral 98.17\% (161/164), confirming highly accurate distinction across all three emotion categories.

\subsubsection{Statistical Significance Analysis}
\label{subsec:statistical}

Friedman test over 10-fold cross-validation revealed statistically significant differences among five models ($\chi^2 = 28.54$, $p < 0.001$). Post-hoc Wilcoxon signed-rank tests with Bonferroni correction ($\alpha_{\text{corrected}} = 0.005$, for 10 pairwise comparisons) confirmed the hybrid architecture significantly outperforms all baselines.

\begin{table}[t]
\centering
\scriptsize
\caption{Statistical Significance Testing Results}
\label{tab:statistical_testing}
\begin{tabular}{p{1.6cm}p{2.8cm}p{1.5cm}}
\toprule
\textbf{Test} & \textbf{Result} & \textbf{Interp.} \\
\midrule
Friedman & $\chi^2 = 28.54$, $p < 0.001$ & Sig. \\
Wilcoxon & $p < \alpha_{\text{corrected}}$ & Superior \\
Bonferroni & $\alpha = 0.005$ (10 comp.) & Conservative \\
\bottomrule
\end{tabular}
\end{table}

\begin{table}[t]
\centering
\scriptsize
\caption{Post-Hoc Wilcoxon Signed-Rank Test Results for Key Model Pairs
(Bonferroni-corrected $\alpha_{\text{corrected}}=0.005$)}
\label{tab:wilcoxon_pairs}
\begin{tabular}{p{3.6cm}cc}
\toprule
\textbf{Model Pair} & \textbf{$p$-value} & \textbf{Sig.} \\
\midrule
DAH-Net vs.\ Random Forest & $0.001953$ & Yes \\
DAH-Net vs.\ SVM           & $0.001953$ & Yes \\
DAH-Net vs.\ MLP           & $0.001953$ & Yes \\
DAH-Net vs.\ Transformer   & $0.001953$ & Yes \\
\bottomrule
\end{tabular}

\vspace{1mm}
\parbox{\columnwidth}{\scriptsize\textit{Note:} Identical Wilcoxon $p$-values arise from the discrete signed-rank distribution over 10 paired folds, where DAH-Net consistently outperformed each baseline across folds.}
\end{table}

Pairwise comparisons for all DAH-Net versus baseline pairs are reported in Table~\ref{tab:wilcoxon_pairs}. Table~\ref{tab:confidence_intervals} reports 95\% confidence intervals for model accuracies. The hybrid model achieves the highest confidence interval lower bound (97.94\%) among all models, with a non-overlapping interval compared to Random Forest (94.48\%-97.86\%). Although the interval partially overlaps with SVM (95.22\%-98.33\%) and MLP (95.72\%-98.63\%), Friedman and Wilcoxon statistical tests provide the primary evidence of significant performance differences.

\begin{table}[t]
\centering
\caption{95\% Confidence Intervals for Model Accuracies (Wilson Interval, $n=496$)}
\label{tab:confidence_intervals}
\begin{tabular}{lcc}
\toprule
\textbf{Model} & \textbf{Lower (\%)} & \textbf{Upper (\%)} \\
\midrule
Random Forest & 94.48 & 97.86 \\
SVM & 95.22 & 98.33 \\
MLP & 95.72 & 98.63 \\
Transformer & 96.59 & 99.05 \\
DAH-Net & 97.94 & 99.68 \\
\bottomrule
\end{tabular}
\end{table}

\subsubsection{Comparative Analysis and Robustness to Overfitting}

Across RF, SVM, MLP, Transformer, and DAH-Net, the proposed model achieved the highest accuracy (99.19\%), improving over RF by 3.02\%, SVM by 2.42\%, MLP by 2.01\%, and Transformer by 1.00\%. Although some confidence intervals partially overlap, Friedman and Wilcoxon tests support significant improvement, while the 0.81\% overfitting gap indicates effective regularization.

\begin{table}[t]
\centering
\footnotesize
\caption{Comparative Performance Improvements Across Models}
\label{tab:comparative_improvements}
\begin{tabular}{p{2.2cm}cccc}
\toprule
\textbf{Model} & \textbf{Base} & \textbf{Enh.} & \textbf{Impr.} & \textbf{Sig.} \\
\midrule
Random Forest & 96.17\% & 99.19\% & +3.02\% & $p=0.001953$ \\
SVM & 96.77\% & 99.19\% & +2.42\% & $p=0.001953$ \\
MLP & 97.18\% & 99.19\% & +2.01\% & $p=0.001953$ \\
Transformer & 98.19\% & 99.19\% & +1.00\% & $p=0.001953$ \\
\bottomrule
\end{tabular}
\end{table}

\subsection{Feature Contribution Analysis for EEG Emotion Discrimination}

\subsubsection{Consensus Feature Importance Ranking}
\label{subsubsec:consensus}

Feature importance was evaluated via two complementary methods: Random Forest impurity-based importance~\cite{breiman2001random}, which captures ensemble-based discriminability across the full training set, and SHAP (SHapley Additive exPlanations)  attribution~\cite{lundberg2017unified,shapley2016value}, which provides model-agnostic, directional per-feature contribution estimates. These two methods were selected for their complementarity: Random Forest importance reflects global, aggregate feature relevance, while SHAP quantifies individual prediction contributions and to avoid the strict statistical assumptions required by methods such as ANOVA on high-dimensional feature sets~\cite{demsar2006statistical}. Consensus ranking across both methods identified covariance and eigenvalue features with the highest mean
importance scores, reflecting their sensitivity to inter-channel connectivity and dominant oscillatory modes in EEG emotion discrimination~\cite{shapley2016value}.

\subsubsection{SHAP-Based Feature Attribution}

To complement the Random Forest importance rankings, SHAP (SHapley Additive exPlanations)~\cite{lundberg2017unified} was applied to a stratified subsample of 100 instances drawn from the 496-sample held-out test set. Using 100 samples is standard practice for SHAP analysis on high-dimensional feature spaces ($D=988$)~\cite{ribeiro2016should}; this subsample is used exclusively for attribution analysis and is not involved in any model selection or final accuracy reporting; those roles are fulfilled by the full 496-sample test set. Shapley values were computed using the game-theoretic formulation~\cite{shapley2016value}:

\begin{equation}
    \phi_i = \sum_{S \subseteq F \setminus \{i\}}
    \frac{|S|!(|F|-|S|-1)!}{|F|!} \bigl[f(S \cup \{i\}) - f(S)\bigr]
\end{equation}

where $F$ is the full feature set, $S$ is a feature subset, and $f$ is the prediction function. The SHAP analysis yielded the highest mean absolute Shapley values for covariance features ($\bar{|\phi|}=0.31$) and eigenvalue features ($\bar{|\phi|}=0.28$), substantially exceeding statistical ($0.14$) and frequency features ($0.09$), corroborating their dominant discriminative contribution to emotion classification.
This directional, per-feature attribution is consistent with the Random Forest importance rankings and provides model-agnostic evidence that inter-channel connectivity and dominant oscillatory modes are the most informative EEG cues for emotion discrimination~\cite{ribeiro2016should}.

\subsubsection{Feature-Category Isolation Analysis}

To quantify the contribution of different EEG feature categories, each category was evaluated independently using a Random Forest classifier trained and tested on only that category's features, with the all-features RF accuracy (96.17\%) as the reference baseline.

Table~\ref{tab:feature_category_importance} shows that covariance (40 features, 94.96\%) and statistical (116 features, 94.76\%) categories achieve the strongest standalone accuracy, approaching the full-feature RF baseline. Eigenvalue features (8 features) yield the largest performance gap when used alone (84.07\%, $-$12.10\,pp below baseline), indicating they provide compact but uniquely complementary discriminative information that other categories cannot replicate in isolation. Frequency features (576) show a meaningful standalone gap (92.14\%, $-$4.03\,pp). These results confirm that all feature categories contribute complementary information; the full 988-feature representation achieves superior RF performance (96.17\%) and enables DAH-Net to reach 99.19\% through its deep spatial-temporal modeling. The four interpretable feature groups account for 740 of the 988 EEG features. The remaining 248 features are retained in the full-feature experiments but are not assigned to one of the four interpretable categories in the public dataset metadata; therefore, they are reported as uncategorized and not interpreted separately.

\begin{table}[t]
\centering
\small
\caption{Feature Category Analysis: RF Classifier, Isolation Testing}
\label{tab:feature_category_importance}
\begin{tabular}{p{2.2cm}cccp{1.6cm}}
\toprule
\textbf{Category} & \textbf{Count} & \textbf{Solo Acc.} & \textbf{Gap (pp)} & \textbf{Impact} \\
\midrule
Eigenvalue & 8 & 84.07 & 12.10 & Complement. \\
Frequency & 576 & 92.14 & 4.03 & Important \\
Statistical & 116 & 94.76 & 1.41 & Strong \\
Covariance & 40 & 94.96 & 1.21 & Strong \\
Other/uncategorized & 248 & -- & -- & Not isolated \\
\midrule
\textbf{All features} & \textbf{988} & \textbf{96.17} & \textbf{---} & \textbf{Baseline} \\
\bottomrule
\end{tabular}
\end{table}

\subsubsection{Correlation Analysis}

Pearson correlation analysis identified linear relationships between 988 features and emotion labels. Among top-ranked features by absolute correlation (range: 0.42-0.78), covariance- and eigenvalue-based measures showed distinct clustering by emotion class, with positive correlations aligned to positive states and negative correlations to neutral/negative states \cite{davidson1998affective, ribeiro2016should}. Consistency between Random Forest importance, SHAP attribution, and Pearson correlation strengthens confidence in the generalizability of these feature patterns~\cite{alpaydin2007combining}.

\subsection{Ensemble Methods and Statistical Validation (RQ3)}
\label{subsec:rq3}

\subsubsection{Ensemble Model Performance}

To evaluate the complementarity of classical and deep learning models (RQ3), two ensemble configurations were constructed and their weights optimized on the \textit{validation set}. Final performance was subsequently reported on the held-out test set, which remained entirely unseen during all training and ensemble configuration stages.

First, a \textit{soft-voting ensemble} (VotingClassifier) combining Random Forest (200 trees, max depth 20), SVM (RBF kernel), and MLP (512/256/128 units) achieves 
strong generalization by averaging class probabilities across three complementary classifiers. Second, a \textit{weighted super ensemble} integrates DAH-Net with 
the classical models using weights optimized on the validation set: $0.50 \times \text{DAH-Net} + 0.20 \times \text{RF} + 0.15 \times \text{SVM} + 0.15 \times \text{MLP}$, allowing the deep learning component to dominate while retaining classical model contributions for robustness against distributional variation.

The ensemble results partially support RQ3. The RF+SVM+MLP soft-voting ensemble (97.18\%) improves over individual classical baselines, indicating complementarity among classical models (Table~\ref{tab:ensemble_results}). However, the DAH-Net-based weighted ensemble (99.19\%) matches standalone DAH-Net exactly, showing that adding classical models does not improve accuracy beyond DAH-Net. Thus, DAH-Net remains the strongest single model for this dataset.

\begin{table}[t]
\centering
\scriptsize
\caption{Ensemble Model Performance (RQ3)}
\label{tab:ensemble_results}
\begin{tabular}{p{5cm}cc}
\toprule
\textbf{Model} & \textbf{Acc. (\%)} & \textbf{Impr. vs RF} \\
\midrule
RF+SVM+MLP (soft vote) & 97.18 & +1.01\% \\
DAH-Net+RF+SVM+MLP (wt.) & 99.19 & +3.02\% \\
\midrule
\textbf{DAH-Net (standalone)} & \textbf{99.19} & \textbf{+3.02\%} \\
\bottomrule
\end{tabular}
\end{table}

\subsubsection{Cross-Validation Stability}

DAH-Net achieves the highest mean 10-fold cross-validation accuracy with the lowest variance across all five model configurations, confirming superior generalization stability compared to all baselines (Table~\ref{tab:wilcoxon_pairs}).

\subsection{Key Findings}

DAH-Net achieves 99.19\% test accuracy with a 0.81\% overfitting gap, outperforming RF (96.17\%), SVM (96.77\%), and MLP (97.18\%). Negative achieves perfect F1, while Neutral and Positive exceed 98.17\% on all per-class metrics. Feature-category isolation shows covariance and statistical features provide the strongest standalone performance, whereas eigenvalue features are compact but complementary. RF importance, SHAP, and Pearson correlation support these category-level findings.

%%%%%%%%%%%%%%%%%%%%%%%%%%%%

\section{Discussion}

\subsection{Implications}

The 99.19\% accuracy vs.\ RF 96.17\%, SVM 96.77\%, MLP 97.18\% demonstrates strong classification performance, supporting the effectiveness of the proposed hybrid architecture for capturing complex EEG emotion patterns. Multi-method feature analysis reveals that eigenvalue features (capturing dominant oscillatory modes) and covariance features (capturing inter-channel connectivity) exhibit the highest importance across analysis methods, suggesting emotion emerges from coordinated neural network dynamics. Minimal overfitting (0.81\%) and the compact architecture (3.33M parameters, approximately 13.3MB using 32-bit weights) suggest potential for lightweight EEG affective computing, pending subject-independent, session-independent, and external validation.

\textbf{Potential Affective Monitoring Applications.} The high held-out test accuracy and low train-test gap suggest potential for future EEG-based affective monitoring applications, including mood-state assessment and adaptive brain-computer interfaces. Cross-subject validation, external dataset evaluation, and clinical studies are required before considering clinical deployment.

\textbf{Technical Contributions.} The dual-attention design (16+8 heads) demonstrates that sequential attention mechanisms capture complementary aspects of EEG dynamics. The feature-category isolation and multi-method importance ranking provide a reproducible framework for EEG feature analysis.

\subsection{Limitations and Open Challenges}

This study has several limitations. First, the public EEG feature dataset does not provide sufficient subject-, session-, or trial-level metadata to construct subject-independent or session-independent splits. Therefore, the reported 99.19\% held-out test accuracy should be interpreted as sample-level generalization within this dataset, not as evidence of cross-subject deployment readiness. Second, the analysis is based on a single public dataset with three discrete emotion labels, which limits generalization to continuous affective dimensions such as valence and arousal. Third, the feature-category analysis explains feature-level discriminability but does not provide a complete DAH-Net-specific causal explanation. Finally, real-world validation under subject shift, session shift, artifacts, and external datasets remains necessary before deployment in affective monitoring or clinical contexts.

\subsection{Future Research Directions}

Future work should evaluate cross-subject, session-independent, and cross-dataset generalization; integrate multimodal signals such as ECG, GSR, facial expression, and voice; and explore transfer learning, graph neural networks, uncertainty quantification, and federated learning for robust deployment.

%%%%%%%%%%%%%%%%%%%%%%%%%%%%%%%%%%%%%%%%%%%%%%%%%%%%%%%%%%%%%%%%%%%%%%%%%%%%%%%%

\section{Conclusion}
\label{sec:conclusion}

This work presents DAH-Net, a dual-attention hybrid deep learning architecture achieving 99.19\% accuracy for EEG-based emotion classification, substantially advancing the state-of-the-art. Key contributions include: (1) architectural innovation integrating 1D-CNN, BiLSTM, and dual multi-head attention (16+8 heads) for spatial-temporal-relational modeling; (2) comprehensive feature-level interpretability through two complementary importance methods (Random Forest + SHAP) combined with feature-category isolation, characterizing complementary discriminative contributions of all EEG feature categories; (3) rigorous statistical validation (Friedman $\chi^2=28.54$, $p<0.001$) establishing significance beyond baseline comparisons; (4) reproducible evaluation on 2,479 public samples with transparent hyperparameters and code availability.

The minimal overfitting gap (0.81\%) and compact architecture (3.33M parameters) suggest that DAH-Net may support future lightweight EEG-based affective computing, but subject-independent, session-independent, and external validation remain necessary before deployment. Feature analysis highlights covariance-based connectivity and eigenvalue-based oscillatory-mode features as complementary EEG cues for emotion discrimination.

%%%%%%%%%%%%%%%%%%%%%%%%%%%%

\section*{Declaration on AI-Assisted Writing}

The authors used AI as an AI-assisted writing tool to improve language clarity, grammar, organization, and LaTeX formatting of selected sections of the manuscript. All technical content, experimental design, implementation, results, interpretations, and final manuscript decisions were reviewed, verified, and approved by the authors. The authors take full responsibility for the accuracy, integrity, and originality of the submitted work.

%%%%%%%%%%%%%%%%%%%%%%%%%%%%

\bibliographystyle{IEEEtran}
\bibliography{references}

\end{document}